\documentclass[runningheads,a4paper]{llncs}

\usepackage{amssymb}
\usepackage{graphicx}

\usepackage{url}
\newcommand{\keywords}[1]{\par\addvspace\baselineskip
\noindent\keywordname\enspace\ignorespaces#1}

\usepackage{graphicx}
\usepackage{cite}

\usepackage{color}
\usepackage{amsmath}
\usepackage{epsfig}
\usepackage{url}

\usepackage{bm}

\newtheorem{thm}{Theorem}
\newtheorem{lem}{Lemma}

\usepackage{amssymb}
\usepackage{amsfonts}

\newcommand{\bsquare}{\hbox{\rule{6pt}{6pt}}}

\newcommand{\bfa}{{\bf a}}

\newcommand{\bff}{{\bf f}}
\newcommand{\bfe}{{\bf e}}
\newcommand{\bfx}{{\bf x}}

\newcommand{\bfr}{{\bf r}}

\def\bsigma{\mbox{\boldmath $\sigma$}}

\newcommand{\R}{{\bf R}}
\newcommand{\A}{{\bf A}}
\newcommand{\B}{{\bf B}}

\newcommand{\I}{{\bf I}}
\newcommand{\D}{{\bf D}}

\newcommand{\X}{{\bf X}}

\newcommand{\bLambda}{{\bf \Lambda}}

\begin{document}

\mainmatter  % start of an individual contribution

\title{Estimation of causal orders in a linear non-Gaussian acyclic model: \\a method robust against latent confounders}

\titlerunning{Estimation of causal orders in LiNGAM with latent confounders}

\author{Tatsuya Tashiro\inst{1}%
%\thanks{Please note that the LNCS Editorial assumes that all authors have used the western naming convention, with given names preceding surnames. This determines the structure of the names in the running heads and the author index.}%
\and Shohei Shimizu\inst{1} \and Aapo Hyv\"arinen\inst{2} \and Takashi Washio\inst{1}}
\authorrunning{T. Tashiro, S. Shimizu, A. Hyv\"arinen and T. Washio}

%\institute{The Institute of Scientific and Industrial Research, Osaka University,\\
%8-1, Mihogaoka, Ibarakishi, Osaka, 567-0047, Japan
\institute{The Institute of Scientific and Industrial Research, Osaka University, Japan
\and Dept. of Mathematics and Statistics, Dept. of Computer Science {\scriptsize /} HIIT, \\University of Helsinki, Finland
%\url{http://www.springer.com/lncs}
}


\toctitle{}
\tocauthor{Tatsuya Tashiro (Osaka University), Shohei Shimizu (Osaka University), Aapo Hyv\"arinen (University of Helsinki), Takashi Washio  (Osaka University)}
\maketitle

\begin{abstract}
We consider to learn a causal ordering of variables in a linear non-Gaussian acyclic model called LiNGAM. 
Several existing methods have been shown to consistently estimate a causal ordering assuming that all the model assumptions are correct. 
But, the estimation results could be distorted if some assumptions actually are violated. 
In this paper, we propose a new algorithm for learning causal orders that is robust against one typical violation of the model assumptions: latent confounders. 
%It enables more accurate estimation of causal orders when some assumption is not very reasonable. 
%In particular, we prove the correctness of our method in the presence of latent confounders. 
%We prove the correctness of our method in the presence of latent confounders and 
We demonstrate the effectiveness of our method using artificial data. 
\keywords{Bayesian networks, causal discovery, non-Gaussianity, latent confounders, independent component analysis}
\end{abstract}

%%%%%%%%
\section{Introduction}
Bayesian networks have been widely used to analyze causal relations of variables in many empirical sciences \cite{Spirtes93book}. 
A common assumption is linear-Gaussianity. 
But this poses serious identifiability problems so that many important models are indistinguishable with no prior knowledge on the structures. 
Recently, it was shown \cite{Shimizu06JMLR} that use of non-Gaussianity allows the full structure of a linear acyclic model to be identified without pre-specifying any causal orders of variables. 
The new model, a Linear Non-Gaussian Acyclic Model called LiNGAM \cite{Shimizu06JMLR}, is closely related to independent component analysis (ICA) \cite{Hyva01book}.

Existing estimation methods \cite{Shimizu06JMLR,Shimizu11JMLR} for LiNGAM learn causal orders assuming that all the model assumptions hold. 
Therefore, these algorithms could return completely wrong estimation results when some of the model assumptions is violated. 
Thus, in this paper, we propose a new algorithm for learning causal orders that is robust against one typical model violation, {\it i.e.}, latent confounders.  
A latent confounder means a variable which is not observed but which exerts a causal influence on some of the observed variables. 
%This leads to a more accurate estimation of these causal orders which do follow the model, e.g., are not confounded by latent confounders. 
% since our algorithm returns only causal orders that fit data well. 
%Specifically, we show that our method works correctly in the presence of latent confounders. 
%We prove the correctness of our method in the presence of latent confounders. 

%The remainder of the paper is organized as follows. 
The paper is organized as follows. 
We first review LiNGAM \cite{Shimizu06JMLR} and its extension to latent confounder cases \cite{Hoyer07IJAR} in Section~\ref{sec:lvlingam}. 
In Section~\ref{sec:est}, we propose a new algorithm to learn causal orders in LiNGAM with latent confounders. 
Simulations are conducted in Section~\ref{sec:exp}. We conclude this paper in Section~\ref{sec:conc}. 
%We empirically evaluate the performance of our algorithm using artificial data in Section~\ref{sec:exp} and conclude this paper in Section~\ref{sec:conc}. 

%%%%%%%%
\section{Background: LiNGAM with latent confounders}\label{sec:lvlingam}
We briefly review a linear non-Gaussian acyclic model called LiNGAM \cite{Shimizu06JMLR} and an extension of the LiNGAM to cases with latent confounding variables \cite{Hoyer07IJAR}. 

In LiNGAM \cite{Shimizu06JMLR}, causal relations of observed variables $x_i$ are modeled as: 
%For simplicity, we present the method using the terminology of the protein network inference cases.
\begin{eqnarray}
x_i &=& \sum_{k(j)<k(i)} b_{ij}x_j + e_i, \label{eq:model1}
\end{eqnarray}
where $k(i)$ is such a causal ordering of variables $x_i$ that they graphically form a directed acyclic graph (DAG) so that no later variable determines, {\it i.e.}, has a directed path on any earlier variable, $e_i$ are external influences, and $b_{ij}$ are connection strengths. 
In matrix form, the model (\ref{eq:model1}) is written as
\begin{eqnarray}
\bfx &=& \B\bfx + \bfe,\label{eq:model2}
\end{eqnarray}
where the connection strength matrix $\B$ collects $b_{ij}$ and the vectors $\bfx$ and $\bfe$ collect $x_i$ and $e_i$.  
Note that the matrix $\B$ can be permuted  to be lower triangular with all zeros on the diagonal if simultaneous equal row and column permutations are made according to a causal ordering $k(i)$ due to the acyclicity.
The zero/non-zero pattern of $b_{ij}$ corresponds to the absence/existence pattern of directed edges. 
External influences $e_i$ follow non-Gaussian continuous distributions with zero mean and non-zero variance and are mutually independent. 
%The independence assumption between $e_i$ means that there is no latent confounding variable. 
The non-Gaussianity assumption on $e_i$ enables identification of a causal ordering $k(i)$ 
based on data $\bfx$ only \cite{Shimizu06JMLR}. 
%without using any background knowledge on the structure \cite{Shimizu06JMLR}. 
This feature is a big advantage over conventional Bayesian networks based on the Gaussianity assumption on $e_i$\cite{Spirtes93book}.

Next, LiNGAM with latent confounders \cite{Hoyer07IJAR} can be formulated as follows:
\begin{eqnarray}
\bfx &=& \B\bfx + \bLambda \bff + \bfe,\label{eq:lvlingam}
\end{eqnarray}
where the difference with LiNGAM~(\ref{eq:model2}) is the existence of latent confounding variable vector $\bff$. 
A latent confounding variable is such an latent  variable that is a parent of more than or equal to two observed variables. 
%$\B$ is a matrix that collects connection strengths $b_{ij}$ from $x_j$ to $x_i$ and can be permuted to be lower triangular, $\bfe$ collects non-Gaussian and independent external influence $e_i$ with zero mean and non-zero variance, 
% as in LiNGAM (\ref{eq:model2}), 
The vector $\bff$ collects non-Gaussian latent confounders $f_j$ with zero mean and non-zero variance $(j=1, \cdots, q)$. Without loss of generality \cite{Hoyer07IJAR}, latent confounders $f_j$ are assumed to be mutually independent. 
The matrix $\bLambda$ collects $\lambda_{ij}$ which denotes the connection strength from $f_{j}$ to $x_{i}$. 
%The matrix $\bLambda$ collects $\lambda_{ij}$ which denotes the connection strength from the $j$-th latent confounder $f_{j}$ to the $i$-th observed variable $x_{i}$. 
For each $j$, at least two $\lambda_{ij}$ are non-zero 
since a latent confounder is defined to have at least two children. 
%since  a latent confounder is defined as a latent variable that has at least two children. 
Further, it is assumed \cite{Hoyer07IJAR} that correlation and conditional correlation of $x_i$, $f_i$ and $e_i$ are entailed by the graph structure only, {\it i.e.}, the zero/non-zero status of $b_{ij}$ and $\lambda_{ij}$. 
This is a well-known assumption called faithfulness in causal discovery \cite{Spirtes93book}. 
%Finally, it is assumed that correlation and conditional correlation of $x_i$, $f_i$ and $e_i$ are entailed by the graph structure, {\it i.e.}, the zero/non-zero status of $b_{ij}$ and $\lambda_{ij}$, but not by special parameter values of $b_{ij}$, $\lambda_{ij}$ and ${\rm std}(e_i)$. This is a well-known assumption called faithfulness in causal discovery \cite{Spirtes93book,Pearl00book}. 

The central problem of causal discovery based on the latent variable LiNGAM in Equation~(\ref{eq:lvlingam}) is to estimate {\it as many} of causal orders $k(i)$ and connection strengths $b_{ij}$ {\it as possible} based on data $\bfx$ only. 
This is because in many cases only an equivalence class of the true model whose members produce the exact same observed distribution is identifiable\cite{Hoyer07IJAR}. 

In \cite{Hoyer07IJAR}, an estimation method based on overcomplete ICA was proposed. 
%:
%But, its application is limited to two or three observed variables since the overcomplete ICA methods are often not very reliable and get stuck in local optima. 
However, overcomplete ICA methods are often not very reliable and get stuck in local optima. 
% as the model. 
% even in the presence of latent confounders. 
%%But, in many cases it cannot uniquely identify the actual model. %The method can be shown to consistently estimate the model in the presence of latent confounders if their number is known.  
%But, in many cases it is not very easy to know their number. 
%But, in many cases it is not very easy to know the number of latent confounders. 
%Further, it is computationally tough and its application is limited to small numbers of observed variables, say 2 or 3. 
%In \cite{Entner10AMBN}, it was proposed to first find unconfounded variable {\it pairs} and then estimates a causal ordering of one to the other instead of a causal ordering of more than three variables. 
Thus, in \cite{Entner10AMBN}, a method that does not use overcomplete ICA was proposed to first find variable {\it pairs} that are not affected by latent confounders and then estimate a causal ordering of one to the other instead of a causal ordering of more than two variables. 
%Thus, in \cite{Entner10AMBN}, a method that does not use overcomplete ICA methods was proposed to first find variable {\it pairs} that are not affected by latent confounders and then estimate a causal ordering of one to the other instead of a causal ordering of more than two variables. 
%%%%%%%%
\section{A hybrid estimation approach}\label{sec:est}
In this section, we propose a new approach for estimating causal orders of more than two variables without explicitly modeling latent confounders. 
We first provide principles to identify such an exogenous (root) variable and a sink variable that are not affected by latent confounders in the latent variable LiNGAM in Equation~(\ref{eq:lvlingam}) (if such variables exist) and next present an estimation algorithm. 
Recent estimation methods \cite{Shimizu11JMLR,Mooij09ICML} for LiNGAM in Equation~(\ref{eq:model2}) and its nonlinear extension \cite{Hoyer09NIPS} learn a causal ordering by finding causal orders one by one either from the top downward or from the bottom upward assuming no latent confounders. 
We extend these ideas to latent confounder cases. 

%\subsection{Estimation principles}\label{sec:identification}
%We first present a principle to find an exogenous variable that has no parent observed variable nor latent confounding variable. 
We first generalize Lemma~1 of \cite{Shimizu11JMLR} for the case of latent confounders. 
\begin{lem}\label{lemma1}
%Assume that the input data $\bfx$ strictly follows the latent variable LiNGAM (\ref{eq:lvlingam}). 
Assume that all the model assumptions of the latent variable LiNGAM (\ref{eq:lvlingam}) are met and the sample size is infinite. 
Denote by $r_i^{(j)}$ the residuals  when $x_i$ are regressed on $x_j$: 
$r_i^{(j)} = x_i - \frac{{\rm cov}(x_i,x_j)}{{\rm var}(x_j)}x_j$ $(i \neq j)$.
Then a variable $x_j$ is an exogenous variable in the sense that it has no parent observed variable nor latent confounder if and only if $x_j$ is independent of its residuals $r_i^{(j)}$ for all $i \neq j$. \mbox{\hfill \qed}
\end{lem} 
%We next present a principle to find a sink variable that has no child observed variable nor latent confounder. 
Next, we generalize the idea of \cite{Mooij09ICML} for the case of latent confounders. 
\begin{lem}\label{lemma2}
%Assume that the input data $\bfx$ strictly follows the latent variable LiNGAM (\ref{eq:lvlingam}). 
%This means that we assume that all the model assumptions are met and the sample size is infinite. 
Assume that all the model assumptions of the latent variable LiNGAM (\ref{eq:lvlingam}) are met and the sample size is infinite. Denote by $\bfx_{(-j)}$ a vector that contains all the variables other than $x_j$.  
Denote by $r_j^{(-j)}$ the residual  when $x_j$ is regressed on $\bfx_{(-j)}$, {\it i.e.}, 
%\begin{eqnarray}
$r_{j}^{(-j)} = x_j - \bsigma^T_{(-j)j} \Sigma_{(-j)}^{-1} \bfx_{(-j)}, $
%\end{eqnarray}
where 
%\begin{eqnarray}
$\Sigma = 
\left[
\begin{array}{cc}
\sigma_{j} & \bsigma_{j(-j)}^T \\
\bsigma_{j(-j)} & \Sigma_{(-j)}
\end{array}
\right]$
%\end{eqnarray}
is the covariance matrix of $[x_j, \bfx_{(-j)}^T]^T$. 
Then a variable $x_j$ is a sink variable in the sense that it has no child observed variable nor latent confounder if and only if $\bfx_{(-j)}$ is independent of its residual $r_j^{(-j)}$. \mbox{\hfill \qed}
\end{lem} 
The proofs of these lemmas are given in the appendix. 
%The proofs of these lemmas are given in the appendix, where Lemma~3 of \cite{Entner10AMBN} is very useful since it has shown that a regressor and its residual are dependent if there are some latent confounders between the regressor and regressand or if there are no latent confounder and the regressand is the ancestor of the regressor in the latent variable LiNGAM (\ref{eq:lvlingam}). 

Thus, we can take a hybrid estimation approach that uses these two principles.  
We first identify an exogenous variable by finding a variable that is most independent of its residuals and remove the effect of the exogenous variable from the other variables by regressing it out. 
We repeat this until independence of every variable and any of its residuals is statistically rejected. 
Dependency between every variable and any of its residuals implies that such an exogenous variable in Lemma~\ref{lemma1} does not exist or some model assumption of latent variable LiNGAM~(\ref{eq:lvlingam}) is violated. 
Similarly, we next identify a sink variable in the remaining variables by finding a variable that its regressors and its residual are most independent and disregard the sink variable. 
We repeat this until independence is statistically rejected for every variable. 
%If some independence assumption is violated, {\it i.e.}, some latent confounder exist, some variable and its residuals are dependent. 
%To evaluate independence between variables and their residuals, we use a kernel-based independence measure called HSIC \cite{GreFukTeoSonSchSmo08}. 
We test pairwise independence between variables and the residuals using a kernel-based independence measure called HSIC \cite{GreFukTeoSonSchSmo08} and combine the resulting $p$-values using a well-known Fisher's method \cite{Fisher50book}. We use Bonferroni correction for multiple comparison dividing the significance level by the maximum number of tests $p$$-$$1$. 

%by the number of causal orders to be estimated $p$$-$$1$. 
%To evaluate independence between variables and their residuals, we use a kernel-based independence measure called HSIC \cite{GreFukTeoSonSchSmo08}. 
%We test independence between variables and their residuals individually and combine the resulting $p$-values using a well-known Fisher's method \cite{Fisher50book}. 

%\subsection{Estimation algorithm}\label{sec:alg}
Thus, the estimation consists of the following steps: 
%We now propose a new algorithm to estimate a causal ordering and the connection strengths in the latent variable LiNGAM~(\ref{eq:lvlingam}):
%
%
%\pagebreak
\noindent
  %\pagebreak
%  \rule{\columnwidth}{0.5mm}
%       { \sffamily
%\vspace{-4mm}
  \begin{enumerate}
  \item Given a $p$-dimensional random vector $\bfx$, a set of its variable subscripts $U$, a $p\times n$ data matrix of the random vector as $\X$ and a significance level $\alpha$, initialize an ordered list of variables $K_{head}:=\emptyset$ and $K_{tail}:=\emptyset$ and $m:=1$. $K_{head}$ and $K_{tail}$ denote first $|K_{head}|$ orders of variables and last $|K_{tail}|$ orders of variables respectively, where each of $|K_{head}|$ and $|K_{tail}|$ denotes the number of elements in the list. 
   \item Let $\Tilde{\bfx}$$=$$\bfx$ and $\Tilde{\X}$$=$$\X$ and find causal orders one by one from the top downward: \label{alg:uppers}
  \begin{enumerate}
   \item Do the following steps for all $j \in U \setminus K_{head}$:  \label{alg:findxm}
 Perform least squares regressions of $\tilde{x}_i$ on $\tilde{x}_j$ for all $i \in U \setminus K_{head}$ $(i \neq j)$ and compute the
residual vectors $\Tilde{\bfr}^{(j)}$. 
%and the residual data matrix $\tilde{\R}^{(j)}$ from the data matrix $\tilde{\X}$. 
Then, find a variable $\tilde{x}_m$ that is most independent of its residuals:
%次式を用いて残差に対して最も独立性の高い $\Tilde{x}_m$ を探す：
     \begin{equation}
      \Tilde{x}_m=\arg \max_{j \in U \setminus K_{head}}P_{Fisher}(\Tilde{x}_j,\Tilde{\bfr}^{(j)}), 
     \end{equation}
     where $P_{Fisher}(\Tilde{x}_j,\Tilde{\bfr}^{(j)})$ is the $p$-value of the test statistic defined as $-2\sum_i \log\{P_{H}(\Tilde{x}_j,\Tilde{r}_i^{(j)})\}$, where $P_{H}(\Tilde{x}_j,\Tilde{r}_i^{(j)})$ is the $p$-value of the HSIC. 
   \item Go to Step~\ref{alg:unders} if $P_{Fisher}(\Tilde{x}_m, \Tilde{\bfr}^{(m)}) < \alpha/(p-1)$. 
   \item \label{step:appendhead} Append $m$ to the end of $K_{head}$ and let $\Tilde{\bfx}:=\Tilde{\bfr}^{(m)}$ and $\Tilde{\X}:=\Tilde{\R}^{(m)}$. 
   If $|K_{head}|=p-1$, append the remaining variable subscript to the end of $K_{head}$ and go to Step~\ref{step:estb}. 
   Otherwise, go back to Step~(\ref{alg:findxm}). 
  \end{enumerate}
 \item If $|K_{head}| < p-2$,\footnote{We do not examine remaining two variables in Step~\ref{alg:unders} since it is already implied in Step~\ref{alg:uppers} that some latent confounders exist or some model assumption is violated.} let $\bfx'=\bfx$ and $\X' =\X$ and $U':=U \setminus K_{head}$ and find causal orders one by one from the bottom upward: \label{alg:unders}
  \begin{enumerate}
   \item Do the following steps for all $j \in U' \setminus K_{tail}$:  \label{alg:findxm2}
 Collect all the variables except $x'_j$ in a vector $\bfx'_{(-j)}$.  Perform least squares regressions of $x'_j$ on $\bfx'_{(-j)}$ and compute the residual ${r'}_j^{(-j)}$.  
% and the residual data matrix from the data matrix $\X'$. 
%    全ての $j \in U' \setminus K_{tail}$ について， $x'_j$ 以外の変数を含むベクトル：$\bfx'_{(-j)}$ で $x'_j$ を最小二乗回帰し，データ行列 $\X'$ から 残差 ${r'}_j^{(-j)}$ を計算する。
    Then, find such a variable $x'_m$ that its regressors and its residual are most independent: 
%    次式を用いて残差に対して最も独立性の高い $x'_m$ を探す：
     \begin{equation}
      x'_m=\arg \max_{j \in U' \setminus K_{tail}}P_{Fisher}(\bfx'_{(-j)},{r'}_j^{(-j)}).
     \end{equation}
  \item Go to Step~\ref{step:estb} if $P_{Fisher}(\bfx'_{(-m)},{r'}_m^{(-m)}) < \alpha/(p-1)$. 
  \item \label{step:append} Append $m$ to the top of $K_{tail}$ and let $\bfx'=\bfx'_{(-m)}$，$\X'=\X'_{(-m)}$. 
  Go to Step~\ref{step:estb} if $|U' \setminus K_{tail}|<3$.\footnotemark[3]
%  \footnote{We do not examine remaining two variables here since it is already implied in Step~\ref{alg:uppers} that some latent confounders exist or some model assumption is violated.} 
  Otherwise go back to Step~(\ref{alg:findxm2}). 
%  Go back to Step~(\ref{alg:findxm2}). 
  \end{enumerate}
% \item 
%% Let $K_{body}:=U \setminus (K_{head} \bigcup K_{tail}) $. 
% {\color{blue} Estimate connection strengths $b_{ij}$ for variables in $K_{head}$ and $K_{tail}$ by regressing each variable $x_i$ in $K_{head}$ and $K_{tail}$ on its non-descendants $x_j$. }
% Let $K:=\{K_{head},K_{body},K_{tail}\}$ as a causal ordering of variables. 
 \item  \label{step:estb} Estimate connection strengths $b_{ij}$ for variables in $K_{head}$ and $K_{tail}$ by doing multiple regression of every variable $x_i$ in $K_{head}$ and $K_{tail}$ on all of its non-descendants $x_j$ with $k(j)<k(i)$. 
  \end{enumerate}
%       } \vspace{-2mm}
%\noindent \rule{\columnwidth}{0.5mm}
%Matlab codes to run the algorithm will be made available on the web. 

%is available at \url{http://www.ar.sanken.osaka-u.ac.jp/~tashiro/xx.html}. 

Note that our algorithm would output no causal orders in cases that such exogenous variables and sink variables as in Lemmas~\ref{lemma1} and \ref{lemma2} do not exist, although the outputs are still correct. 
One way to learn more causal orders in those cases would be to develop a divide-and-conquer algorithm that divides variables into subsets where such exogenous or sink variables exist and integrates the estimation results on the subsets. 
This is an important direction of future research. 

\section{Experiments on artificial data}\label{sec:exp}
%We compare our method with some existing methods \cite{Shimizu06JMLR,Shimizu11JMLR,Entner10AMBN} on artificial data and simulated fMRI data. 

\begin{figure}[tb]
  \begin{center}
   \includegraphics[width=0.55\textwidth]{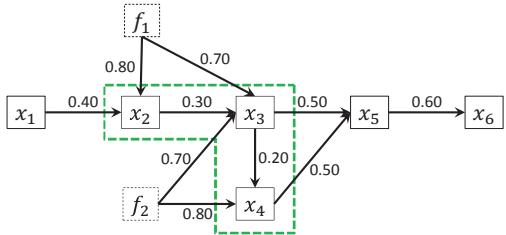}
   \caption{True network used in the simulation. The variables $f_1$ and $f_2$ were latent confounders. The green contours include variables that share $f_1$ or $f_2$. The external influences $e_i$ are omitted to be shown.}
   \label{fig:sim_truedag}
  \end{center}
\end{figure}

%%%%%%%%
%\subsection{Experiments on artificial data}
%As a sanity check, we conducted experiments on artificial data to evaluate the performance of our method. 
%in estimating causal orders $k(i)$. 
%For comparison, we also tested an estimation method for LiNGAM called DirectLiNGAM \cite{Shimizu11JMLR}. 
%If there is no latent confounders, Both of our method and DirectLiNGAM should estimate correct causal orders for enough large sample sizes. 
%For comparison, we also tested an estimation method for LiNGAM called DirectLiNGAM \cite{Shimizu11JMLR} and a method for LiNGAM with latent confounders called Pairwise~LvLiNGAM \cite{Entner10AMBN}. 
We compared our method with an estimation method for LiNGAM~(\ref{eq:model2}) called DirectLiNGAM \cite{Shimizu11JMLR} that does not allow latent confounders and an estimation method for latent variable LiNGAM~(\ref{eq:lvlingam}) called Pairwise~LvLiNGAM \cite{Entner10AMBN}. 
If there is no latent confounders, all the methods should estimate correct causal orders for enough large sample sizes. 
The number of variables was 6, and the sample sizes tested were 500, 1000, 2000. 
The original network used was shown in Figure~\ref{fig:sim_truedag}.  
The $e_1$, $e_4$ and $f_1$ followed a multimodal asymmetric mixture of two Gaussians, 
$e_2, e_5, f_2$ followed a double exponential distribution, 
and $e_3$ and $e_6$ followed a multimodal symmetric mixture of two Gaussians. 
%The standard deviations of the external influences were set so that their signal-to-noise ratios were all ones. 
The standard deviations of the $e_i$ were set so that their signal-to-noise ratios, {\it i.e.}, ${\rm var}(x_i)/{\rm var}(e_i)$$-$$1$ were all ones. 
%{\color{blue} (分布なに?)}
%Here, $K_{head}$$=$$x_1$ and $K_{tail}$$=$$x_5,x_6$. 
The number of trials was 100. The significance level $\alpha$ was 0.05. 

First, to evaluate performance of estimating causal orders $k(i)$, we computed the percentage of correctly estimated causal orders in estimated causal orders between two variables (Precision) and the percentage of correctly estimated causal orders in actual causal orders between two variables that share no latent confounders in the true data generating network (Recall). 
The reason why only pairwise causal orders were evaluated was that Pairwise LvLiNGAM only estimates causal orders of two variables unlike our method and DirectLiNGAM. 
% for our method and DirectLiNGAM \cite{Shimizu11JMLR}. 
Tables~\ref{Precision} and \ref{Recall} show the results. 
%Fig.~\ref{fig:sim_estdag} shows an estimated network by our method after pruning by a sparse regression method called Bolasso \cite{Bach08ICML}, 
%which was randomly chosen from the estimated networks in the condition with sample size 1000. 
Regarding precisions, our method was comparable to Pairwise LvLiNGAM and the two methods were much better than DirectLiNGAM for all the conditions. 
Regarding recalls, our method was better than both DirectLiNGAM and Pairwise LvLiNGAM for all the conditions. 
%For both precisions and recalls, our method and Pairwise LvLiNGAM were much better than DirectLiNGAM for all the conditions. 
%For recalls, DirectLiNGAM was better than our method for all the conditions except the case with sample size 500. 
%This shows that causal orders estimated by our method were much more reliable than DirectLiNGAM that does not examine the existence of latent confounders. 

%On the other hand, our method was more conservative. 
%To visualize the estimation results,  
%Next, Figure~\ref{fig:plot} gives scatterplots of the estimated connection strengths $b_{ij}$ of our method and DirectLiNGAM versus the true ones. Note that Pairwise~LvLiNGAM does not estimate $b_{ij}$. 
%The different plots correspond to different numbers of variables and different sample sizes, where each plot combines the data for different adjacency matrices B and 18 different distributions of the external influences p(ei). 
Next, to evaluate the performance in estimating connection strengths $b_{ij}$, 
we computed the root mean square errors between true connection strengths and estimated ones. 
The root mean square errors for our method and DirectLiNGAM were 0.079 and 0.090 for 500 data points, 0.070 and 0.079 for 1000 data points and 0.015 and 0.057 for 2000 data points, respectively, where our method was more accurate. 
Note that Pairwise~LvLiNGAM does not estimate $b_{ij}$. 
%We can see that our method worked better than DirectLiNGAM, as evidenced by the grouping of the data points onto the main diagonal.

\begin{table}[tb]
\begin{minipage}{0.46\textwidth}
\begin{center}
\caption{Precisions}
 \begin{tabular}{l|ccc}
%\multicolumn{4}{c}{Precision} \\
    & \multicolumn{3}{c}{Sample size}  \\
    & 500     & 1000     & 2000     \\ \hline 
%   Our method  & \textcolor{red}{0.87} & \textcolor{red}{1.00} & \textcolor{red}{1.00} \\ 
%  DirectLiNGAM  & 0.63     & 0.82 & 0.83     
   Our method  & 0.78 & 0.80 & 0.80 \\ 
  DirectLiNGAM  & 0.65     & 0.64 & 0.64  \\
  Pairwise LvLiNGAM & 0.79 & 0.81 & 0.81
 \end{tabular}
\label{Precision}
\end{center}
\end{minipage}
\hspace*{7mm}
\begin{minipage}{0.46\textwidth}
\begin{center}
\caption{Recalls}
 \begin{tabular}{l|ccc}
%\multicolumn{4}{c}{Recall} \\
    & \multicolumn{3}{c}{Sample size} \\
    & 500     & 1000     & 2000     \\ \hline 
%0.3 &    Our method  & \textcolor{red}{0.79} & 0.82 & 0.83 \\ 
%  & DirectLiNGAM  & 0.76     & \textcolor{red}{0.98} & \textcolor{red}{1.00}     \\
%   &  Our method  & \textcolor{red}{0.79} & 0.82 & 0.83 \\ 
%  & DirectLiNGAM  & 0.76     & \textcolor{red}{0.98} & \textcolor{red}{1.00}     \\
%   &  Our method  & \textcolor{red}{0.79} & 0.82 & 0.83 \\ 
%  & DirectLiNGAM  & 0.76     & \textcolor{red}{0.98} & \textcolor{red}{1.00}     
  Our method  & 0.97 & 0.99 & 0.99 \\ 
 DirectLiNGAM  & 0.81     & 0.80 & 0.81  \\
 Pairwise LvLiNGAM & 0.86 & 0.89 & 0.90
 \end{tabular}
\label{Recall}
\end{center}
\end{minipage}

\end{table}%

\section{Conclusions}\label{sec:conc}
We proposed a new algorithm for learning causal orders, which is robust against latent confounders. 
%We discussed a linear non-Gaussian acyclic model with latent confounders and proposed a new algorithm for learning  causal orders with examining existence of latent confounders. 
In experiments on artificial data, our approach learned more causal orders accurately than two existing methods. In future work, we would like to test our method on real-world data including functional magnetic resonance imaging data to analyze causal interactions between brain regions.

\subsubsection*{Acknowledgments.}
%This work was supported by KAKENHI \#21700302 and \#22300054.
S.S and T.W. were supported by KAKENHI \#24700275 and \#22300054. 
We thank Patrik Hoyer and Doris Entner for helpful comments. 
%S.S and T.W. were supported by MEXT Grant-in-Aid for Young Scientists \#21700302 and JSPS Grant-in-Aid for Scientific Research (B) \#22300054, respectively.

\bibliographystyle{splncs03} 
\bibliography{shimizu12a,tsuika_bun}

%%%%%%%%
\section*{Appendix: Proofs of the lemmas}\label{appendix}
%We first quote Darmois-Skitovitch (D-S) theorem \cite{Darmois1953,Skitovitch53}:  
%We first quote Darmois-Skitovitch theorem \cite{Darmois1953,Skitovitch53} and Lemma~3 of \cite{Entner10AMBN}. 
%\begin{thm}[Darmois-Skitovitch theorem (D-S theorem) \cite{Darmois1953,Skitovitch53}]\label{thm1}
\begin{thm}[Darmois-Skitovitch theorem (D-S theorem) \cite{Darmois1953}]\label{thm1}
Define two random variables $y_1$ and $y_2$ as linear combinations of independent random variables $s_i$($i$$=$1, $\cdots$, $q$): $y_1 = \sum_{i=1}^q \alpha_is_i$, $y_2 = \sum_{i=1}^q \beta_i s_i$.
%\begin{eqnarray}
%y_1 = \sum_{i=1}^q \alpha_is_i, \hspace{2mm}y_2 = \sum_{i=1}^q \beta_i s_i.\nonumber
%\end{eqnarray}
Then, if $y_1$ and $y_2$ are independent, all variables $s_j$ for which $\alpha_j\beta_j\neq0$ are Gaussian. \mbox{\hfill \qed}
\end{thm}
In other words, this theorem means that if there exists a non-Gaussian $s_j$ for which $\alpha_j\beta_j$$\neq$$0$, $y_1$ and $y_2$ are dependent. 

Further, Lemma~3 of \cite{Entner10AMBN} has shown that the regressor and its residual in simple linear regression are dependent if there are some latent confounders between the regressor and regressand in the latent variable LiNGAM (\ref{eq:lvlingam}).

%\begin{lemma}[Lemma~3 of \cite{Entner10AMBN} ]\label{entner}
%Define two random variables $y_1$ and $y_2$ as linear combinations of independent random variables $s_i$($i$$=$1, $\cdots$, $q$): $y_1 = \sum_{i=1}^q \alpha_is_i$, $y_2 = \sum_{i=1}^q \beta_i s_i$.
%%\begin{eqnarray}
%%y_1 = \sum_{i=1}^q \alpha_is_i, \hspace{2mm}y_2 = \sum_{i=1}^q \beta_i s_i.\nonumber
%%\end{eqnarray}
%Then, if $y_1$ and $y_2$ are independent, all variables $s_j$ for which $\alpha_j\beta_j\neq0$ are Gaussian. \mbox{\hfill \qed}
%\end{lemma}

\subsubsection*{Proof of Lemma~\ref{lemma1}}
i) Assume that $x_j$ has at least one parent observed variable or latent confounder. 
%i) Assume that $x_j$ is not an exogenous variable in the sense that it has at least one parent observed variable or latent confounder. 
%Let a vector $\x_{P_j}$ and a column vector $\w_{P_j}$ collect all the variables in $P_j$ and the corresponding connection strengths, respectively. 
%Then, the covariances between $\x_{P_j}$ and $x_j$ are 
%%\begin{eqnarray}
%$E(\x_{P_j}x_j) = E\{\x_{P_j} (\w_{P_j}^T \x_{P_j}+e_j) \} 
%% &=& E(\x_{P_j}\w_{P_j}^T \x_{P_j})+E( \x_{P_j}e_j) \nonumber \\
% = E(\x_{P_j} \x_{P_j}^T)\w_{P_j}.  $ %\label{eq:cov}
%%\end{eqnarray}
%The covariance matrix $E(\x_{P_j} \x_{P_j}^T)$ is positive definite since the external influences and latent confounders are mutually independent and have  positive variances.
%% The covariance matrix $E(\x_{P_j} \x_{P_j}^T)$ is positive definite since the external influences $e_h$ that correspond to those parent variables $x_h$ in $P_j$ are mutually independent and have  positive variances.
%Thus, the covariance vector $E(\x_{P_j}x_j)=E(\x_{P_j} \x_{P_j}^T)\w_{P_j}$ above cannot equal the zero vector, 
%and there must be at least one variable in $P_j$ with which $x_j$ covaries. 
%%, that is, {\rm cov}$(x_i,x_j)$$\neq$$0$. 
Let $P_j$ denote the set of the parent variables of $x_j$. 
Then one can write $x_j$$=$$ \sum_{p_h \in P_j} w_{jh} p_h$$+$$e_j$, where the parent variables $p_h$ are independent of $e_j$ and the coefficients $w_{jh}$ are non-zero. 
Suppose that $x_i$ is a parent of $x_j$. 
For such $x_i$, we have 
%Then, for such a variable $x_i$, we have 
%\begin{eqnarray}
$r_i^{(j)} = x_i - \frac{{\rm cov}( x_i, x_j )}{{\rm var}(x_j)}x_j  = x_i - \frac{{\rm cov}( x_i, x_j )}{{\rm var}(x_j)}(\sum_{p_h \in P_j} w_{jh} p_h + e_j) 
 = \left\{1-\frac{w_{ji}{\rm cov}(x_i,x_j)}{{\rm var}(x_j)}\right\} x_i -\frac{{\rm cov}(x_i,x_j)}{{\rm var}(x_j)}\sum_{p_h\in P_j,  p_h\neq x_i}w_{jh}p_h 
    - \frac{{\rm cov}(x_i,x_j)}{{\rm var}(x_j)} e_j. $
% &=& \left\{1-\frac{b_{ji}{\rm cov}(x_i,x_j)}{{\rm var}(x_j)}\right\} x_i - \frac{{\rm cov}(x_i,x_j)}{{\rm var}(x_j)} e_j \nonumber \\
%  & & -\frac{{\rm cov}(x_i,x_j)}{{\rm var}(x_j)}\sum_{x_h\in P_j, h\neq i}b_{jh}x_h. \\
%\end{eqnarray}
Each of those parent variables (including $x_i$) in $P_j$ is a linear combination of external influences {\it other than} $e_j$ and latent confounders that are non-Gaussian and independent. 
%due to the relation of $x_h$ to $e_j$ that $x_j = \sum_{h \in P_j} b_{jh} x_h + e_j = \sum_{h \in P_j} b_{jh} \left(\sum_{k(t)\le k(h)} a_{ht}e_t\right) + e_j$ , where $e_t$ and $e_j$ are independent. 
%Each of those parent variables $x_i$ and $x_h$ in $P_j$ is a linear combination of external influences other than $e_j$ due to Equation~(\ref{eq:ica}) and the equation $x_j = \sum_{h \in P_j} b_{jh} x_h + e_j$. %, where $x_h$ and $e_j$ are independent. 
Thus, the $r_i^{(j)}$ and $x_j$ can be written as linear combinations of non-Gaussian and independent external influences including $e_j$ and latent confounders.
Further, the coefficient of $e_j$ on $r_i^{(j)}$  is non-zero since ${\rm cov}(x_i,x_j)\neq 0$ due to the faithfulness and that on $x_j$ is one by definition.  
These imply that $r_i^{(j)}$ and $x_j$ are dependent since $r_{i}^{(j)}$, $x_j$ and $e_j$ correspond to $y_1$, $y_2$, $s_j$ in D-S theorem, respectively.  
Next, for the other case that $x_j$ has a latent confounder,  $r_{i}^{(-j)}$ and an observed  variable can be shown to be dependent using Lemma~3 of \cite{Entner10AMBN} since by definition at least one observed variable shares the latent confounder with $x_j$. 

 ii) The converse of contrapositive of i) is straightforward using the model definition.  
From i) and ii), the lemma is proven. 
\mbox{\hfill \bsquare}

\vspace{-3mm}

\subsubsection*{ Proof of Lemma~\ref{lemma2}}
i) Assume that a variable $x_j$  has at least one child observed variable or latent confounder. 
%i) Assume that a variable $x_j$ is not a sink variable in the sense that it has at least one child observed variable or latent confounder. 
%Based on a similar argument as in Lemma~\ref{lemma1},  $\bsigma^T_{(-j)j} \Sigma_{(-j)}^{-1}$ in 
%$r_{j}^{(-j)} = x_j - \bsigma^T_{(-j)j} \Sigma_{(-j)}^{-1} \bfx_{(-j)}$ cannot be the zero vector. 
%\end{eqnarray}
First, without loss of generality, one can write
\begin{eqnarray}
\bfx = 
\left[
\begin{array}{c}
x_j\\
\bfx_{(-j)}
\end{array}
\right]
&=& (\I-\B)^{-1} (\bLambda \bff + \bfe)
= \A (\bLambda \bff + \bfe)\\
%&=&
%\left[
%\begin{array}{cc}
%1 & \bfa_{j(-j)}^T\\
%\bfa_{(-j)j} & \A_{(-j)}
%\end{array}
%\right]
%\left[
%\begin{array}{c}
%d_j\\
%\bfd_{(-j)}
%\end{array}
%\right]\\
&=&
\left[
\begin{array}{cc}
1 & \bfa_{j(-j)}^T\\
\bfa_{(-j)j} & \A_{(-j)}
\end{array}
\right]
\left[
\begin{array}{c}
{\pmb \lambda}_{j}^T\bff+e_j\\
\bLambda_{(-j)}\bff+\bfe_{(-j)}
\end{array}
\right], 
\end{eqnarray}
where each of $\A$ ($=(\I-\B)^{-1}$) and $\A_{(-j)}$ is invertible and can be permuted to be a lower triangular matrix with the diagonal elements being ones if the rows and columns are simultaneously permuted according to the causal ordering $k(i)$. The same applies to the inverse of $\A$: 
%The inverse of $\A$, {\it i.e.}, $\I-\B$, can also be permuted to be a lower triangular matrix with the diagonal element being ones: 
\begin{eqnarray}
\A^{-1} &=& 
\left[
\begin{array}{cc}
(1-\bfa_{j(-j)}^T\A_{(-j)}^{-1}\bfa_{(-j)j})^{-1} & -\bfa_{j(-j)}^T\D^{-1} \\
-\D^{-1}\bfa_{(-j)j} & \D^{-1} 
\end{array}
\right],
\end{eqnarray}
where $\D=\A_{(-j)}-\bfa_{(-j)j}\bfa_{j(-j)}^T$.
Thus, $1-\bfa_{j(-j)}^T\A_{(-j)}^{-1}\bfa_{(-j)j}=1$. 
Then, 
%we have 
\begin{eqnarray}
r_{j}^{(-j)} &=& x_j - \bsigma^T_{(-j)j} \Sigma_{(-j)}^{-1} \bfx_{(-j)}\\
 &=& {\pmb \lambda}_j^T\bff + e_j + \bfa^T_{j(-j)}(\bLambda_{(-j)}\bff + \bfe_{(-j)})\nonumber \\
 & &- \bsigma^T_{(-j)j} \Sigma_{(-j)}^{-1} \{\bfa_{(-j)j}({\pmb \lambda}_{j}^T\bff+e_j)+\A_{(-j)}(\bLambda_{(-j)}\bff+\bfe_{(-j)})\\
&=& \{{\pmb \lambda}_j^T+\bfa_{j(-j)}^T\bLambda_{(-j)}- \bsigma^T_{(-j)j} \Sigma_{(-j)}^{-1}(\bfa_{(-j)j}{\pmb \lambda}_j^T + \A_{(-j)}\bLambda_{(-j)})\}\bff \nonumber\\
& & \hspace{-2.5mm} +\{1- \bsigma^T_{(-j)j} \Sigma_{(-j)}^{-1} \bfa_{(-j)j}\}e_j  + \{ \bfa_{j(-j)}^T - \bsigma^T_{(-j)j} \Sigma_{(-j)}^{-1} \A_{(-j)} \}\bfe_{(-j)}. \label{eq:prf_r}
\end{eqnarray}
In Equation~(\ref{eq:prf_r}), if $\bfa_{j(-j)}^T - \bsigma^T_{(-j)j} \Sigma_{(-j)}^{-1} \A_{(-j)} = {\bf 0}^T$, then we have
%In Equation~(\ref{eq:prf_r}), if $\bfa_{j(-j)}^T - \bsigma^T_{(-j)j} \bSigma_{(-j)}^{-1} \A_{(-j)}$ is a zero vector, that is, $\bfa_{j(-j)}^T \A_{(-j)}^{-1} = \bsigma^T_{(-j)j} \bSigma_{(-j)}^{-1}$, then we have
\begin{eqnarray}
r_{j}^{(-j)} &=& \{{\pmb \lambda}_j^T(1-\bfa_{j(-j)}^T \A_{(-j)}^{-1}\bfa_{(-j)j})\}\bff
 +\{1- \bfa_{j(-j)}^T \A_{(-j)}^{-1} \bfa_{(-j)j}\}e_j \\
 &=& {\pmb \lambda}_j^T\bff +e_j. 
\end{eqnarray}
Thus, the coefficient of $e_j$ on $r_{j}^{(-j)}$ is one. 
Now, suppose that $x_j$ has a child $x_i$.  
%Then, at least one variable in $\bfx_{(-j)}$ is a child of $x_j$. 
%Let us denote by $x_i$ such a child of $x_j$. 
The coefficient of $e_j$ on $x_i$ is non-zero due to the faithfulness.  
Thus, $r_{j}^{(-j)}$ and $x_i$ are dependent due to D-S theorem. 
Next, suppose that $x_j$ has a latent confounder $f_i$.
Then, in Equation~(\ref{eq:prf_r}), the corresponding element in ${\pmb \lambda}_j$ is not zero, {\it i.e.}, the coefficient of $f_i$ on $r_j^{(-j)}$ is not zero. 
Further, $f_i$ has a non-zero coefficient on at least one variable in $\bfx_{(-j)}$ due to the definition of latent confounders and faithfulness. 
Therefore, $r_{j}^{(-j)}$ and $\bfx_{(-j)}$ are dependent due to D-S theorem. 

On the other hand, in Equation~(\ref{eq:prf_r}), if $\bfa_{j(-j)}^T - \bsigma^T_{(-j)j} \Sigma_{(-j)}^{-1} \A_{(-j)} \neq {\bf 0}^T$, at least one of the coefficients of the elements in $\bfe_{(-j)}$ on $r_{j}^{(-j)}$ is not zero. 
By definition, every element in $\bfe_{(-j)}$ has a non-zero coefficient on the corresponding element in $\bfx_{(-j)}$, 
Thus, $r_{j}^{(-j)}$ and $\bfx_{(-j)}$ are dependent due to D-S theorem.

 ii) The converse of contrapositive of i) is straightforward using the model definition. 
From i) and ii), the lemma is proven. 
\mbox{\hfill \bsquare}

\end{document}